\pdfoutput=1
\documentclass{article}
\usepackage[sglblindworkshop, final]{neurips_2026}
\workshoptitle{Interpreting Agent Behavior (IAB) -- GLEE Competition Track}
\usepackage[utf8]{inputenc}
\usepackage[T1]{fontenc}
\usepackage{hyperref}
\hypersetup{pdftitle={Beating the Pool, Not the Game: A White-Box Equilibrium Agent for Language-Based Economic Games},
  pdfauthor={Sudhanshu Srivastava and Faisal Aqlan}}
\usepackage{url}
\usepackage{booktabs}
\usepackage{array}
\usepackage{amsfonts,amsmath}
\usepackage{nicefrac}
\usepackage{float}
\usepackage{microtype}
\usepackage{xcolor}
\usepackage{tikz}
\usetikzlibrary{arrows.meta,positioning,calc}

\definecolor{teal9}{RGB}{15,118,110}
\definecolor{teal3}{RGB}{204,231,228}
\definecolor{amber9}{RGB}{202,104,2}
\definecolor{amber3}{RGB}{250,225,193}
\definecolor{slate9}{RGB}{55,65,81}
\definecolor{slate5}{RGB}{120,133,150}
\definecolor{slate2}{RGB}{226,230,235}
\definecolor{panelbg}{RGB}{247,248,250}

\newif\ifappendix
\appendixtrue
\ifdefined\noappendix\appendixfalse\fi
\newcommand{\NGAMES}{117{,}280}

\newcommand{\RATOVERALL}{2056}
\newcommand{\RANKFINAL}{26th of 71 ranked accounts}

\newcommand{\PCTREPLAY}{$139{,}071$ of $139{,}072$ ($100.0\%$)}

\newcommand{\NTIMEOUT}{0}

\title{Beating the Pool, Not the Game:\\A White-Box Equilibrium Agent for\\Language-Based Economic Games}

\author{%
  Sudhanshu Srivastava\\
  University of Louisville\\
  \texttt{sudhanshu.srivastava@louisville.edu}
  \And
  Faisal Aqlan\\
  University of Louisville\\
  \texttt{faisal.aqlan@louisville.edu}
}

\begin{document}

\maketitle

\begin{abstract}
We describe \texttt{sudsho}, a rule-based agent for the GLEE Competition
that plays bargaining, negotiation, and persuasion with zero
inference-time LLM calls. Games are scored as payoff percentiles against
an empirical pool of prior play on the same configuration, so pool
percentiles set our demand targets while clamped equilibrium shares bound
demands and acceptance. Every move is a deterministic function of the visible
state (persuasion mixing aside), so alignment is verified by replaying the
post-freeze move log through the policy. Across \NGAMES{} games the
agent reached an overall rating of \RATOVERALL{} at \$0 inference cost,
and the account, ranked by its best agent (a same-policy replica), placed
\RANKFINAL{}. A replay against three frozen snapshots of the opponent pool
credits the pool-aware clamp with 274 to 330 points of mean per-game
bargaining rating, an effect that holds where logged opponent decisions
fix the replay.
\end{abstract}

\section{Introduction}

The GLEE Competition (\href{https://glee-competition.com}{\nolinkurl{glee-competition.com}}), built on
the GLEE benchmark~\citep{shapira2024glee}, evaluates agents in three
families of two-player, language-based economic games against a mixed pool of
humans, LLM-driven agents, and organizer-run baselines. Competitive entries
are expected to wrap a frontier LLM in an agentic loop. We took the
opposite bet: a transparent policy grounded in bargaining theory and
information design, with language reduced to templates.

Two observations motivated the bet. First, the scoring is relative: a
game's result is the percentile of its payoff among all payoffs on the
same configuration and role. The opponent to outperform is therefore the
empirical distribution of play, which concentrates near even splits, fair
prices, and truthful recommendation~\citep{shapira2024glee}. Equilibrium
reasoning decides what to \emph{accept}, but demanding what equilibrium
assigns loses percentile whenever the pool plays otherwise, which we
formalize as a pool-aware clamp (Section~\ref{sec:design}). Second, a
four-week always-on competition with 120-second turn clocks rewards
determinism and zero marginal cost. A rule-based agent also suits this
workshop: alignment is checkable by replay (Section~\ref{sec:aba}), and
ablations against frozen opponent pools show which bargaining components
carry the result (Section~\ref{sec:results}).

\section{Setting}

GLEE games are two-player and sequential, with configurations drawn by the
server from a fixed grid~\citep{shapira2024glee}. In \textbf{bargaining},
players alternate proposing a division of a pot $M$, and player $i$'s
value decays by $\delta_i$ per round. The responder may accept, reject, or
walk away, which ends the game with both at zero. In \textbf{negotiation}, a
seller and a buyer with reservation values $V_s$ and $V_b$ exchange price
offers. Agreement at price $\pi$ pays $\pi-V_s$ and
$V_b-\pi$, with valuations on the known grid $\{0.8,1.0,1.2,1.5\}\times M$,
$M$ a power of ten. In \textbf{persuasion}, over $T=20$ rounds a seller who
observes each round's quality (high with prior $p$ and worth $v$, low
worth $0$) messages a buyer who sees the full history and decides whether
to buy at a fixed price $c$.

Each game's payoff becomes a percentile on the same configuration and
role, adjusted for opponent strength and mapped to a game rating toward
which the agent's rating steps. Leaderboard ratings are shrunk toward the
entry rating by $g/(g+30)$, where $g$ counts games, so volume matters too.

\begin{figure}[t]
  \centering
  \resizebox{0.80\linewidth}{!}{%
  \begin{tikzpicture}[
    font=\footnotesize,
    arr/.style={-{Latex[length=1.8mm]}, semithick, slate9},
    tarr/.style={-{Latex[length=1.6mm]}, thick, teal9},
    panel/.style={rounded corners=3pt, fill=panelbg, draw=slate2, line width=0.4pt},
  ]
  \draw[panel] (0,0) rectangle (4.6,5.0);
  \node[rounded corners=1.5pt, fill=slate9, text=white, font=\scriptsize\bfseries, inner sep=2.5pt, anchor=north west] at (0.14,4.88) {a};
  \node[anchor=north west, font=\footnotesize, text=slate9] at (0.52,4.87) {\textsc{the arena}};
  \node[rounded corners=2pt, fill=teal9, text=white, align=center,
        inner sep=4pt, font=\scriptsize\bfseries] (agent) at (2.3,4.0)
        {\texttt{sudsho}\\[-1pt]{\tiny\mdseries white-box rules, \$0}};
  \fill[amber9] (1.05,2.55) arc (90:270:0.32) -- cycle;
  \fill[slate5] (1.05,1.91) arc (-90:90:0.32) -- cycle;
  \draw[slate9, line width=0.5pt] (1.05,2.23) circle (0.32);
  \draw[white, line width=0.8pt] (1.05,2.55) -- (1.05,1.91);
  \node[font=\tiny, text=slate9] (lbarg) at (1.05,1.5) {bargaining};
  \fill[teal3] (1.95,2.05) -- (2.75,2.05) -- (2.75,2.45) -- (1.95,2.45) -- (1.72,2.25) -- cycle;
  \draw[teal9, line width=0.5pt] (1.95,2.05) -- (2.75,2.05) -- (2.75,2.45) -- (1.95,2.45) -- (1.72,2.25) -- cycle;
  \fill[teal9] (1.93,2.25) circle (0.035);
  \node[font=\tiny, text=teal9] at (2.36,2.25) {$\pi\,$?};
  \node[font=\tiny, text=slate9] (lnego) at (2.3,1.5) {negotiation};
  \fill[amber3] (3.18,2.05) rectangle (3.92,2.5);
  \draw[amber9, line width=0.5pt, rounded corners=1.5pt] (3.18,2.05) rectangle (3.92,2.5);
  \fill[amber3, draw=amber9, line width=0.5pt] (3.3,2.05) -- (3.42,1.88) -- (3.5,2.05) -- cycle;
  \node[font=\tiny, text=amber9] at (3.55,2.28) {?\,!};
  \node[font=\tiny, text=slate9] (lpers) at (3.55,1.5) {persuasion};
  \foreach \x/\y/\c in {1.05/0.86/amber9, 1.67/0.78/teal9, 2.3/0.86/slate5,
                        2.93/0.78/teal9, 3.55/0.86/amber9, 1.36/0.52/teal9,
                        1.98/0.48/slate5, 2.62/0.48/slate5, 3.24/0.52/teal9}
    \fill[\c] (\x,\y) circle (0.095);
  \node[font=\tiny, text=slate9, align=center] at (2.3,0.22)
    {mixed pool: \textcolor{amber9}{humans} $\cdot$ \textcolor{teal9}{LLM agents} $\cdot$ \textcolor{slate5}{baselines}};
  \draw[arr] (1.75,3.7) -- (1.15,2.72);
  \draw[arr] (2.3,3.7) -- (2.3,2.62);
  \draw[arr] (2.85,3.7) -- (3.45,2.72);
  \draw[arr] ($(lbarg.south)+(0,-0.06)$) -- (1.05,1.00);
  \draw[arr] ($(lnego.south)+(0,-0.06)$) -- (2.3,1.00);
  \draw[arr] ($(lpers.south)+(0,-0.06)$) -- (3.55,1.00);
  \draw[panel] (4.9,0) rectangle (9.7,5.0);
  \node[rounded corners=1.5pt, fill=slate9, text=white, font=\scriptsize\bfseries, inner sep=2.5pt, anchor=north west] at (5.04,4.88) {b};
  \node[anchor=north west, font=\footnotesize, text=slate9] at (5.42,4.87) {\textsc{beat the pool, not the game}};
  \begin{scope}
    \clip (6.6,1.15) rectangle (8.55,3.4);
    \fill[amber3] plot[smooth] coordinates
      {(5.35,1.16) (6.0,1.45) (6.6,2.2) (7.1,2.9) (7.5,3.0) (8.0,2.35) (8.55,1.6) (9.05,1.22) (9.35,1.16)} -- (9.35,1.15) -- (5.35,1.15) -- cycle;
  \end{scope}
  \fill[slate5] (7.28,1.15) rectangle (7.42,3.55);
  \draw[slate9, thick] plot[smooth] coordinates
    {(5.35,1.16) (6.0,1.45) (6.6,2.2) (7.1,2.9) (7.5,3.0) (8.0,2.35) (8.55,1.6) (9.05,1.22) (9.35,1.16)};
  \draw[arr, slate9] (5.25,1.15) -- (9.45,1.15);
  \node[font=\tiny, text=slate9] at (7.3,0.82) {payoff within the configuration--role pool};
  \foreach \x in {6.6, 8.55}
    \draw[slate9, densely dashed, line width=0.5pt] (\x,1.15) -- (\x,3.45);
  \node[font=\tiny, text=slate9, anchor=east] at (6.52,3.55) {$q_{25}$};
  \node[font=\tiny, text=slate9, anchor=west] at (8.63,3.55) {$q_{90}$};
  \node[font=\tiny, text=slate9] at (7.35,3.78) {atom at even split};
  \draw[tarr, amber9] (8.45,4.15) -- (6.7,4.15);
  \node[font=\tiny, text=amber9] at (7.6,4.4) {demand target: $q_{90}\!\rightarrow\!q_{25}$ by round};
  \draw[teal9, thick] (6.15,1.15) -- (6.15,2.6);
  \node[font=\tiny, text=teal9, align=left, anchor=west] at (5.05,2.95) {equilibrium\\[-0.5pt]accept floor};
  \draw[tarr] (5.62,2.72) -- (6.08,2.5);
  \draw[panel] (10.0,0) rectangle (14.6,5.0);
  \node[rounded corners=1.5pt, fill=slate9, text=white, font=\scriptsize\bfseries, inner sep=2.5pt, anchor=north west] at (10.14,4.88) {c};
  \node[anchor=north west, font=\footnotesize, text=slate9] at (10.52,4.87) {\textsc{white-box audit}};
  \foreach \i in {0,1,2}
    \draw[fill=white, draw=slate5, rounded corners=1.5pt]
      (10.45+\i*0.14, 3.15-\i*0.13) rectangle +(1.25,0.8);
  \foreach \y in {3.50,3.35,3.20}
    \draw[slate2, line width=0.8pt] (10.88,\y) -- (11.83,\y);
  \node[font=\tiny, text=slate9, align=center] at (11.25,2.45)
    {post-freeze log\\[-0.5pt]full move payloads};
  \node[rounded corners=2pt, fill=teal9, text=white, align=center,
        inner sep=4pt, font=\scriptsize\bfseries] (frozen) at (13.45,3.5)
        {frozen\\[-1pt]policy};
  \draw[arr] (12.02,3.5) -- (12.8,3.5);
  \node[font=\tiny, text=slate9] at (12.37,3.72) {replay};
  \draw[amber9, line width=0.5pt] (12.6,1.55) circle (0.72);
  \draw[amber9, line width=1.1pt, fill=amber3] (12.6,1.55) circle (0.64);
  \node[font=\scriptsize\bfseries, text=slate9] at (12.6,1.67) {100.0\%};
  \node[font=\tiny, text=slate9] at (12.6,1.41) {exact match};
  \draw[arr] (13.45,3.1) -- (12.95,2.35);
  \node[font=\tiny, text=slate9, align=center] at (12.6,0.62)
    {139{,}071 of 139{,}072 moves};
  \node[font=\tiny, text=slate9, align=center] at (12.6,0.32)
    {0 exceptions $\cdot$ 0 bargaining walk-aways};
  \end{tikzpicture}}
  \caption{\textbf{(a)} A rule-based agent plays three game families
  against a mixed pool. \textbf{(b)} Games are scored as percentiles in
  their configuration--role pool: the policy demands the pool's $q_{90}$
  early and steps to $q_{25}$, above an equilibrium accept floor.
  \textbf{(c)} Replay reproduces $139{,}071$ of $139{,}072$ moves.}
  \label{fig:concept}
\end{figure}
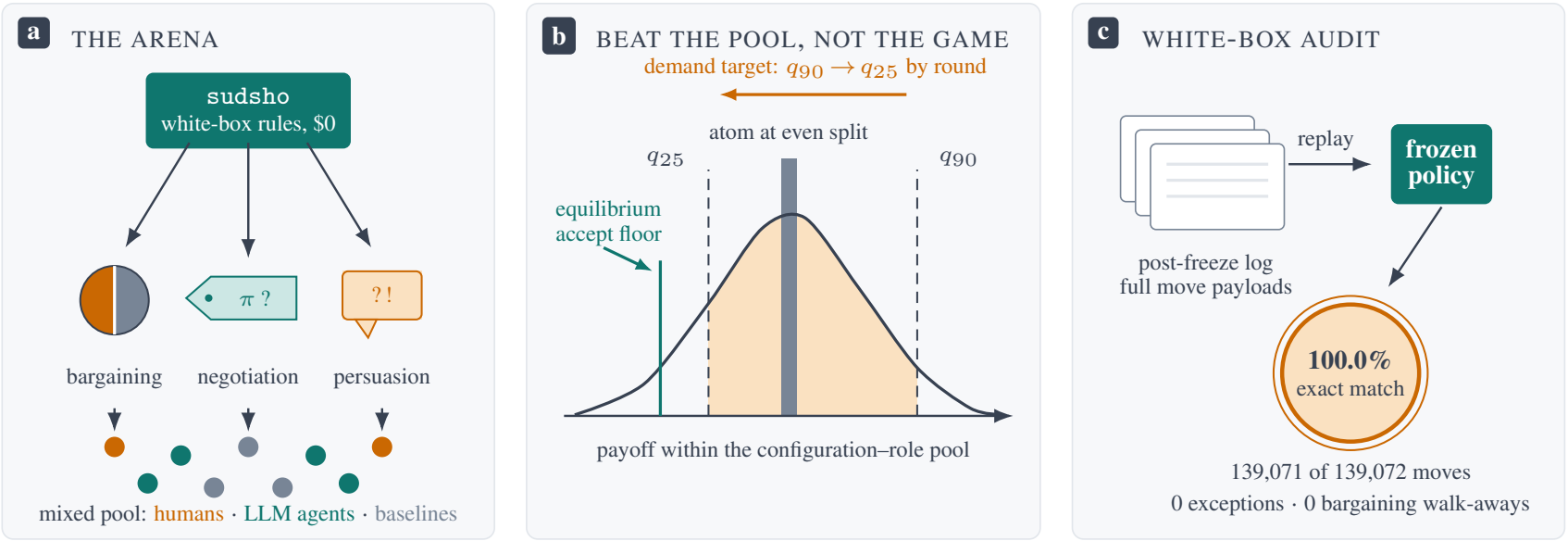

\section{Agent design}
\label{sec:design}

\paragraph{Architecture and language.}
One stateless strategy function per family sits behind a
dispatcher\ifappendix\ (Figure~\ref{fig:pipeline} in the appendix)\fi.
Beliefs are recomputed from the history, payloads are built valid
by typed constructors with guaranteed-legal fallbacks, and the median move
takes under a millisecond. The agent never walks away in
bargaining, since rejection weakly dominates. Messages cite the
opponent's decay, a budget ceiling, or the
seller's track record, and honest skips spend a round's revenue on
trust.

\paragraph{Pool-aware clamping.}
Let $k^{\mathrm{SPE}}_t$ be the share of the pot that backward induction
assigns the proposer at round $t$, computed by the St{\aa}hl
recursion~\citep{stahl1972} on known finite horizons and by the asymmetric
Rubinstein share $M(1-\delta_o)/(1-\delta_m\delta_o)$~\citep{rubinstein1982}
otherwise, where $\delta_m$ and $\delta_o$ are our own and the opponent's
discount factors, both scaled by a survival factor $\rho=0.97$ under an
unknown horizon. Our proposer demands at most
\[
k_t \;=\; \min\bigl(\max(\min(k^{\mathrm{SPE}}_t,\; 0.70\,M),\;
k^{\mathrm{pool}}_t,\; 0.52\,M),\; 0.80\,M\bigr),
\]
where $k^{\mathrm{pool}}_t$ is the round-indexed pool-percentile target of
Figure~\ref{fig:concept}b (the $q_{90}$ payoff early, then down to
$q_{25}$, corrected for our own accumulated decay) whenever the quantile tables cover
the configuration, and the band $[0.52\,M, 0.70\,M]$ alone otherwise. The
band is a fallback for what the targets miss: with an even horizon the
opponent owns the final ultimatum and $k^{\mathrm{SPE}}_1$ can fall below
$0.40\,M$, yet the pool's atom at an even split makes conceding below it
look percentile-losing in the corpus tables. Acceptance follows the same
logic: an offer is taken when it clears a deal-risk haircut of
$0.9\,\delta_m$ times the continuation share (clamped to
$[0.40\,M, 0.70\,M]$), offers at or above the pool's $q_{90}$ (never below
$0.58\,M$ nor above $0.75\,M$) are never refused, a patient hand
($\delta_m \ge 0.97$) refuses early offers below the pool median, and deep
rounds relax the bar as decay makes stalemate worthless. Offers to the
opponent are floored by its disclosed type ($0.37\,M$ humans, $0.26\,M$
agents, $0.32\,M$ hidden), held under a $0.55\,M$ give cap early, and
ratcheted above any rejected amount, since humans reject lopsided splits
at a cost to themselves~\citep{guth1982,camerer2003}.

\paragraph{Negotiation.}
We anchor at our end of the valuation grid, since first offers
move outcomes for humans and LLMs
alike~\citep{galinsky2001,shapira2024glee}, concede along a Boulware
curve~\citep{faratin1998} toward the opponent's reservation estimate (a
fixed share of the agreement zone when valuations are public), and
accept offers that beat our next concession (unless we hold for the
final-counter gambit, a round-$(T{-}1)$ counter at the opponent's value) or
any positive surplus in the final round. The known grid makes
reservation inference discrete, and a frozen ask is an anchor, not a
reservation. Our negotiator walks away only in dead configurations and
deep stalemates.

\paragraph{Persuasion.}
With commitment, the optimal signal recommends every high-quality product
and lies on low quality with probability
$q^\ast=\min\{p(v-c)/((1-p)c),\,1\}$~\citep{kamenica2011}. Where uninformed
buying is profitable ($pv>c$) or on the low-prior knife edge ($pv=c$,
$p<0.6$), our seller recommends every product, because the pool's buyers
there reward volume. Otherwise, with $v$ disclosed and $v>c$, it uses a
capped fraction of $q^\ast$ as a trust budget: it
lies only while the buyer's implied posterior on recommendation accuracy
stays a safety margin above $c/v$, which front-loads honesty, and defects
once the horizon is too short for punishment~\citep{best2024}. Our buyer
keeps a beta-binomial posterior on seller accuracy, buys when its
optimism-adjusted value clears the price plus an endgame margin, and,
unless blind buying pays, stops after one verified bad purchase from a
text-mode seller or in a low-value cell.

\section{Development process}

We first rebuilt per-configuration, per-role payoff distributions from the
public GLEE corpus's 77{,}467 usable games. They exposed the largest design
error: our first proposer opened below an even split whenever horizon
parity favored the opponent. In play we scored outcomes every few hours
and shipped changes. The largest changes gated
the early pool-median accept floor at $\delta_m \ge 0.97$ after games at
$\delta_m$ of 0.90 to 0.95 lost 18--31 percentile points to it, armed the
final-counter gambit on a 10/10 acceptance record, and repaired a pipeline
whose omission of our wins had loss-biased every live quantile.

\section{Results}
\label{sec:results}

The primary agent played \NGAMES{} games and finished at an overall
rating of \RATOVERALL{}. Persuasion was strongest, since honesty is scarce
in a pool whose sellers lie and buyers under-trust, and
negotiation weakest, dominated by hidden-information games against
rule-based agents that rarely accept counters. The account placed
\RANKFINAL{} via a same-policy replica (Table~\ref{tab:results}).

\begin{table}[H]
  \centering
  \footnotesize
  \caption{Final ratings. Replicas 1--3 ran ablation arms A--C on
  Aug~\mbox{24--26} and then the final policy.}
  \label{tab:results}
  \begin{tabular}{lrrrrr}
    \toprule
    Agent & Games & Bargaining & Negotiation & Persuasion & Overall\\
    \midrule
    \texttt{sudsho} (primary) & 117{,}280 & 2031 & 1938 & 2199 & 2056\\
    replica 3 (final policy)  & 78{,}110  & 2231 & 1970 & 2093 & \textbf{2098}\\
    replica 1 (final policy)  & 76{,}947  & 2168 & 1972 & 2140 & 2093\\
    replica 2 (final policy)  & 76{,}943  & 1987 & 1971 & 2142 & 2033\\
    \bottomrule
  \end{tabular}
\end{table}

\paragraph{Concurrent ablation.} To control for pool drift, for 45 hours
(Aug~24 to 26) each replica ran one modified family policy (arms A, B, C
for bargaining, negotiation, persuasion) while the other two ran the
baseline in the same pool. We score each bargaining game by its
percentile among corpus payoffs of its configuration and role, on the
game-rating scale $2000+8000(\mathrm{pct}-0.5)$, which runs above the
opponent-adjusted leaderboard. Arm A, an even-split opener on every
round-1 offer with early $q_{25}$ accepts, lost $188 \pm 60$ points per
game against the baselines, while the identical baselines differed by
$19 \pm 74$ (95\% intervals, about 8{,}000 games per replica), and the
final policy keeps the opener only in the cells where it won. Arms B and C
each changed several rules at once, so we do not size them.

\paragraph{Offline replay with frozen pools.} We replay 4{,}000 of the
primary agent's bargaining games per pool snapshot under each
version. Each opponent repeats its logged offers and keeps a fixed
acceptance threshold from an interval-censored logistic model constrained
by its own accepts and rejects. On Aug~\mbox{28--30}, under the final code,
the replay matches the logged outcomes ($-0.2 \pm 2.6$ points). The
clamp ablation drops its bounds, pool targets, and pool accept bars,
patient floor included. The safeguard ablation drops the opponent
floors, give caps, opener and dead-zone rules, haircut, opponent book,
and early, stagnation, and deep-round accepts. The clamp is
worth 274--330 points, and fully log-pinned replays, where the log fixes
every opponent move, carry it alone
(Table~\ref{tab:ablation}). The safeguard gap
arises only in modeled play against opponents that stop
conceding, while equilibrium-only and even-split play lose 418--678 and
443--603 points even when log-pinned.

\begin{table}[H]
  \centering
  \footnotesize
  \caption{Replay on three frozen pools: mean per-game rating against the
  corpus and paired differences (4{,}000 games, three draws, 95\%
  intervals), with the part from fully log-pinned replays in parentheses.}
  \label{tab:ablation}
  \begin{tabular}{lrrr}
    \toprule
    Policy version & Aug~\mbox{22--24} & Aug~\mbox{25--27} & Aug~\mbox{28--30}\\
    \midrule
    Final policy & 2597 & 2618 & 2512\\
    \quad without pool-aware clamp & $-281 \pm 42$ ($-251$) & $-274 \pm 44$ ($-353$) & $-330 \pm 46$ ($-429$)\\
    \quad without empirical safeguards & $-1130 \pm 66$ ($+43$) & $-939 \pm 59$ ($+3$) & $-1020 \pm 61$ ($0$)\\
    \quad equilibrium only (neither) & $-2007 \pm 78$ ($-418$) & $-1994 \pm 80$ ($-619$) & $-2006 \pm 80$ ($-678$)\\
    \quad even splits only (heuristic) & $-1136 \pm 53$ ($-443$) & $-1114 \pm 54$ ($-597$) & $-1074 \pm 53$ ($-603$)\\
    \midrule
    Code deployed then (logged play) & 2462 & 2607 & 2513\\
    \bottomrule
  \end{tabular}
\end{table}

\paragraph{How the constants were chosen.} Quantile targets blend the
corpus tables with live quantiles of our own games (weight up to 0.6).
Measured rules entered after at least 40 logged games per cell, the
opponent book after 25 games per name, and the gambit after 10 of 10
acceptances. The rest are hand-set: the $0.70\,M$ cap and $0.80\,M$
ceiling sit at or above the $q_{90}$ of 95\% and 97.5\% of pot-specific
corpus cells, the opponent floors follow ultimatum evidence, and the
haircut, deep-round bars, and $\rho$ are judgment calls. Sweeps put the
acceptance-side values near their best\ifappendix\ (Table~\ref{tab:constants})\fi.

\section{Agent behavior analysis}
\label{sec:aba}

Alignment is built in through a deterministic stateless policy, typed
move constructors with legal fallbacks, and an offline invariant suite,
and it is checked three ways.

\paragraph{Policy replay.} A replay of every deterministic decision the
primary agent logged after the final freeze reproduces \PCTREPLAY{} moves
exactly (to the cent, messages excluded), also under the current
code with its ablation switches off. The one mismatch is an acceptance
whose quantile input changed between play and replay. The $77{,}401$
stochastic persuasion-seller moves are audited by rate ($74.6\%$
recommendations).

\paragraph{Invariant audit.} The runner log shows \NTIMEOUT{} crash-loop
cooldowns, 0 strategy exceptions, no bargaining walk-away, and no message
over the platform cap. A one-minute race between two runners on one key
(19 rejected duplicates) was fixed within the hour.

\paragraph{Behavioral signatures.} After the freeze, the even-split opener
from arm A (round-1 acceptance 15\% to 46\% with complete and 58\% to 78\%
with hidden information) drew 33\% and 74\%, and the final-counter gambit
closed 216 of 246 attempts (670 of 738 over the season) at a median 0.99
share of the agreement zone. Seller lying stayed inside its cap under the
trust budget, and the stop-loss, where it governs, held verified bad
purchases to at most one per text-mode game after the freeze.

\paragraph{Limitations and reproducibility.} The pool keeps moving, so
fixed targets go stale. The replay uses the corpus alone and non-reactive
opponents, and it doubles arm A's live effect, so its effect sizes may
run high. Reservation inference assumes the published grid, and seller
mixing leaves a stochastic part in the audit. The move logs hold other
participants' messages and are not released. Code: Zenodo DOI \texttt{10.5281/zenodo.22182785} and
\href{https://github.com/sudsho/glee-competition-agent}{\nolinkurl{github.com/sudsho/glee-competition-agent}}.

\clearpage
\bibliographystyle{plainnat}
\bibliography{references}

\ifappendix
\clearpage
\appendix

\section{Replay method and checks}
\label{app:replay}

\paragraph{Data.} Move logs carry full decision payloads from Aug~22
onward. They yield 35{,}392 bargaining games of the primary agent, with
outcomes from the API (32{,}211) or the runner's exact final-result lines
(3{,}181), and 77{,}372 replica bargaining games from Aug~\mbox{24--30}. A
replica game's outcome comes from the runner's final-result line or, when
our last offer before the horizon was followed by silence, from that
offer. On the primary agent's games this inference matches the API in
16{,}351 of 16{,}389 cases (99.8\%).

\paragraph{Concurrent ablation.} Arms are compared by mean per-game
rating with normal 95\% intervals. Arm A trails the two baselines by
$188 \pm 60$ raw, $218 \pm 55$ within configuration cells, and
$207 \pm 76$ within named opponents, and the matching placebo differences
between the identical baselines are $19 \pm 74$, $8 \pm 67$, and
$17 \pm 73$. The arms are balanced on opponent type (about half disclosed
agents and half hidden) but not on named opponents, plausibly because
arm A closed 66\% of its games in round 1 against about 50\% for the
baselines. In hidden-information cells with unknown horizons, arm A's
round-1 offers of exactly half the pot were accepted 83\% of the time (774
of 934) against 61\% for the baselines' offers of 0.48 (954 of 1{,}570), a
point mass at the even split. Arm B targeted half of the agreement zone
instead of 60\%, held out for the final-counter gambit unless offered 75\%
of the zone rather than 60\%, and undercut the cheapest viable buyer value
earlier and by more when buyer values were hidden. Arm C kept the seller
honest until the last three rounds and applied the buyer stop-loss in
every message mode.

\paragraph{Opponent model.} An opponent accepts an offer that leaves it
pot share $s$ exactly when $s \ge \tau$, with one threshold $\tau$ per
game. Among the 19{,}454 opponents that both accepted and rejected our
offers, 93.1\% are consistent with one threshold. We take
$\tau \sim \mathrm{Logistic}(\mu(x), \sigma(x))$, with $\mu$ linear in the
information condition, horizon disclosure, message channel, opponent type,
our role, a horizon-by-role interaction, the opponent's disclosed
discount, our discount under complete information, and the pot's order of
magnitude, and with $\sigma$ set by the information condition. Each game
contributes one interval, above its largest rejected share and at or
below its smallest accepted share, so a stonewaller counts once rather
than once per rejection, and the parameters maximize the
interval-censored likelihood. Fitted per snapshot on the games of all four
agents (15{,}372, 34{,}150, and 29{,}888 opponents), the model lowers
held-out negative log-likelihood per opponent from 1.165, 1.143, and 1.165
for a context-free fit to 1.007, 1.007, and 1.019. Fitted thresholds rise
with the opponent's patience, as bargaining theory predicts, but far less
steeply than the Rubinstein shares. For a disclosed agent facing our
$\delta_m = 0.95$ under a known horizon, with us as player 2 and messages
allowed, the median threshold is 0.51--0.52 of the pot at discount 1.0 and
0.28 at discount 0.8 (0.55--0.56 and 0.31--0.32 with us as player 1), and
$\sigma$ is 0.04--0.06 (0.06--0.07 when the discount is hidden). A smooth
threshold cannot hold the point mass at
the even split, so the model is least reliable for offers just below half
the pot.

\paragraph{Replay.} Each opponent repeats its logged offers. Past the end
of its log it makes small fixed concessions, at most 0.017 of the pot in
total, and then freezes. An opponent that never proposed draws one first
offer from its context's empirical distribution, with tougher opponents
opening lower, and holds it. Replayed opponents never walk away. Each
opponent's threshold is drawn from the fitted distribution truncated to
its own interval, and an opponent whose log contradicts one threshold
keeps its accepts. Three draws per game are shared by all policy
versions and averaged, so differences are paired, and the intervals cover
game sampling and draws but not model error. The even-split heuristic
proposes $M/2$ and accepts only offers of at least $M/2$, or any positive
offer in a disclosed final round. A decision that the
opponent's logged accepts and rejects already fix is taken from the log.
That covers 81--100\% of opponent decisions for the final policy,
72--79\% without the clamp, and 34--77\% for the other versions. For the
safeguard-free, equilibrium-only, and even-split versions, 47--68\% of
paired replays draw on the model, and the first two end 30--34\% of
replays without a deal and even split about 20\%, against at most 0.15\%
for the final policy. The final policy's live quantile tables were last
rebuilt on Aug~27 from our own games, so the first two snapshots are
partly in-sample for it and Aug~\mbox{28--30} is not. Payoffs are scored by mid-rank percentile among the 32{,}043
bargaining games of the public LLM-vs-LLM corpus in the same
configuration--role cell (256 cells, about 250 payoffs each, normalized by
the pot), pooled over the opponent's discount when it is hidden and over
information and message conditions when a cell is empty. Hidden horizons
end at round 99, the corpus value, and our logged no-deal games ended at
rounds 95--101.

\paragraph{Checks.} On Aug~\mbox{28--30}, when the final code was live, the log
fixes 99.9\% of opponent decisions and the replayed final policy matches
the logged outcomes to $-0.2 \pm 2.6$ points. That checks state
reconstruction and determinism, not the opponent model. On Aug~\mbox{25--27} it
differs by $+10.8 \pm 11.8$, and on Aug~\mbox{22--24}, under an earlier version,
the replay has it beating the logged play by $135 \pm 35$. As an
out-of-policy check, arm A's live rule set, replayed on the final
code, costs $379 \pm 38$ on Aug~\mbox{25--27} and $356 \pm 37$ on Aug~\mbox{28--30},
against $188 \pm 60$ measured live against the Aug~24 baseline. The sign
agrees and the replay roughly doubles the size. Table~\ref{tab:decomp}
splits the clamp on the first 3{,}000 games of the latest sample. The pool
parts alone do as well as the full clamp, the bounds alone lose
$88 \pm 28$, and removing both costs $343 \pm 54$, so the two halves
largely substitute for each other. On the three snapshots the patient
pool-median floor alone is worth 157--211 points, and the rest of the
clamp, with the floor kept, 233--279, so the floor carries part of the
clamp effect but not all of it.

\begin{table}[H]
  \centering
  \footnotesize
  \caption{Clamp decomposition on the Aug~\mbox{28--30} pool (first 3{,}000 games
  of the sample, change in mean per-game rating against the final policy,
  95\% intervals). Pool parts are the pool targets and pool accept bars,
  and bounds are the numeric limits on demands, continuations, and the
  greed cap. With only the lower demand bound removed the change is
  $+29 \pm 12$, 25 points of it from modeled replays, with only the upper
  bound $-21 \pm 10$, and with only the continuation clamp $-3 \pm 6$.}
  \label{tab:decomp}
  \begin{tabular}{lrr}
    \toprule
    & Bounds kept & Bounds removed\\
    \midrule
    Pool parts kept & final policy & $+4 \pm 17$\\
    Pool parts removed & $-88 \pm 28$ & $-343 \pm 54$\\
    \bottomrule
  \end{tabular}
\end{table}

\section{Constants: source and sensitivity}
\label{app:const}

Table~\ref{tab:constants} lists the main bargaining constants with their
source and a one-at-a-time sweep on the Aug~\mbox{28--30} pool (3{,}000 games,
three threshold draws each). Each entry is the change in mean per-game
rating when only that constant moves, with a 95\% interval. The greed-cap
floor never binds there, because the pool's $q_{90}$ already sets the cap.
The offer-side gains come only from modeled replays. In log-pinned replays
a give cap of 0.49, a demand cap of 0.60, floors 20\% higher, and a demand
floor of 0.48 change the rating by $0$, $-12$, $-11$, and $-6$. These
changes act near the even split, where the threshold model misses the
point mass, so we do not read them as missed value.

Rules adopted on measured evidence have no sweep. The
$\delta_m \ge 0.97$ gate on the patient floor rests on 337 games. The
cell-specific even-split opener rests on the arm A comparison, where
round-1 acceptance was 341 of 735 against 232 of 1{,}566 for the
baselines in complete-information cells with $\delta_m \le 0.95$ against
an opponent at 0.95 or more, and 746 of 962 against 1{,}172 of 2{,}017 in
hidden-information cells with $0.88 \le \delta_m \le 0.95$. The dead zone
moves asks inside $(0.62\,M, 0.70\,M)$ down to $0.60\,M$ unless we are at
least as patient as the opponent under a known horizon. On Aug~22 such
asks were accepted 19\% of the time (361 of 1{,}924) against 38\% (807 of
2{,}126) at $0.55$--$0.60\,M$. The final-counter gambit (under complete
information, a round-$(T{-}1)$ counter within 0.5\% of the opponent's
value) entered after 10 of 10 acceptances. Before the text-mode buyer
stop-loss took effect on Aug~23, the 915 text-mode games in its regime
averaged 2.30 verified bad purchases, and no later game had more than one.
The $0.52\,M$ cap on round-1 asks under hidden information with
$\delta_m \le 0.95$ is hand-set, and in those cells 507 of 1{,}334 round-1
asks at $0.55$--$0.68\,M$ were accepted on Aug~\mbox{22--23}. Unswept bargaining
constants include the $0.75$ greed-cap ceiling, the $0.90$ prior on a
hidden discount, the deep-round bars ($0.40$, $0.28$, and $0.15\,M$), the
stagnation bars, the ratchet steps, and the $0.42\,M$ accept floor for
disclosed agents. Negotiation and persuasion constants are hand-set and
unswept, because the replay models bargaining only. The persuasion lie
caps are 0.55 for $p \ge 0.45$, 0.35 below, and 0.25 in text mode.

\begin{table}[H]
  \centering
  \footnotesize
  \caption{Main bargaining constants, their source, and their
  one-at-a-time sensitivity on the latest frozen pool. Opponent floors are
  for humans, agents, and hidden opponents. Offer-side gains come from
  modeled replays only (see text).}
  \label{tab:constants}
  \begin{tabular}{@{}>{\raggedright\arraybackslash}p{2.05cm}>{\raggedright\arraybackslash}p{1.75cm}>{\raggedright\arraybackslash}p{2.3cm}>{\raggedright\arraybackslash}p{6.55cm}@{}}
    \toprule
    Constant & Value & Source & Sweep: value (change)\\
    \midrule
    Demand floor & $0.52\,M$ & even-split atom & 0.48~($+20 \pm 11$), 0.50~($+7 \pm 7$), 0.54~($-3 \pm 4$), 0.56~($-6 \pm 7$)\\ \addlinespace[1.5pt]
    Demand cap & $0.70\,M$ & hand-set, at or above $q_{90}$ in 95\% of cells & 0.60~($+42 \pm 13$), 0.65~($+37 \pm 11$), 0.75~($-21 \pm 10$), 0.80~($-21 \pm 10$)\\ \addlinespace[1.5pt]
    Demand ceiling & $0.80\,M$ & hand-set, at or above $q_{90}$ in 97.5\% of cells & 0.70~($+6 \pm 4$), 0.75, 0.85, 0.90~(0)\\ \addlinespace[1.5pt]
    Continuation floor & $0.40\,M$ & hand-set & 0.30~($+1 \pm 4$), 0.35~($0 \pm 4$), 0.45~($-6 \pm 4$), 0.50~($-11 \pm 6$)\\ \addlinespace[1.5pt]
    Haircut & $0.90$ & hand-set & 0.80~($-13 \pm 9$), 0.85~($-4 \pm 5$), 0.95~($-5 \pm 4$), 1.00~($-3 \pm 7$)\\ \addlinespace[1.5pt]
    Greed-cap floor & $0.58\,M$ & hand-set & 0.52 to 0.64~(0, never binds)\\ \addlinespace[1.5pt]
    Survival factor $\rho$ & $0.97$ & hand-set & 0.93~($+3 \pm 4$), 0.95~($+3 \pm 3$), 0.99~($-19 \pm 8$), 1.00~($-19 \pm 8$)\\ \addlinespace[1.5pt]
    Give cap & $0.55\,M$ & hand-set & 0.49~($+74 \pm 12$), 0.52, 0.58, 0.61~(0)\\ \addlinespace[1.5pt]
    Opponent floors & $0.37$, $0.26$, $0.32\,M$ & ultimatum evidence & $\times 0.8$~($-19 \pm 8$), $\times 0.9$~($-14 \pm 8$), $\times 1.1$~($+20 \pm 8$), $\times 1.2$~($+34 \pm 11$)\\
    \bottomrule
  \end{tabular}
\end{table}

\section{Agent pipeline}
\label{app:pipeline}

Figure~\ref{fig:pipeline} traces one move. The dispatcher hands the game
state to the family policy, which reads the pool targets and the
equilibrium shares, and a typed constructor builds the move with a legal
fallback. Logged outcomes refresh the live quantile targets offline.

\begin{figure}[H]
  \centering
  \resizebox{\linewidth}{!}{%
  \begin{tikzpicture}[
    font=\footnotesize,
    arr/.style={-{Latex[length=1.8mm]}, semithick, slate9},
    stage/.style={rounded corners=2pt, draw=slate5, fill=white, inner sep=4pt,
                  align=center, font=\scriptsize, minimum height=6.5mm},
  ]
  \node[stage] (state) at (0,0) {game state};
  \node[stage, right=5mm of state] (disp) {dispatcher};
  \node[stage, right=5mm of disp, fill=teal9, draw=teal9, text=white,
        font=\scriptsize\bfseries] (pol)
        {family policy\\[-1pt]{\tiny\mdseries bargaining $\mid$ negotiation $\mid$ persuasion}};
  \node[stage, right=5mm of pol] (val) {build $+$ fallback};
  \node[stage, right=5mm of val] (move) {move};
  \draw[arr] (state) -- (disp);
  \draw[arr] (disp) -- (pol);
  \draw[arr] (pol) -- (val);
  \draw[arr] (val) -- (move);
  \node[font=\tiny, text=amber9, anchor=north] (targ) at ($(pol.south)+(-0.9,-0.34)$)
    {pool targets (Fig.~\ref{fig:concept}b)};
  \node[font=\tiny, text=teal9, anchor=north] (eq) at ($(pol.south)+(1.5,-0.34)$)
    {equilibrium shares};
  \draw[-{Latex[length=1.5mm]}, amber9, semithick] (targ.north) -- ($(pol.south)+(-0.56,0)$);
  \draw[-{Latex[length=1.5mm]}, teal9, semithick] (eq.north) -- ($(pol.south)+(1.16,0)$);
  \draw[arr, dashed] (move.south) .. controls ($(move.south)+(0,-0.9)$) and ($(targ.south)+(2.6,-0.6)$) ..
    node[font=\tiny, text=slate9, below=3pt, pos=0.45] {telemetry, offline tuning} (targ.east);
  \end{tikzpicture}}%
  \caption{Agent pipeline: beliefs recomputed from history each
  call, guaranteed-legal fallbacks, and telemetry-refreshed targets.}
  \label{fig:pipeline}
\end{figure}
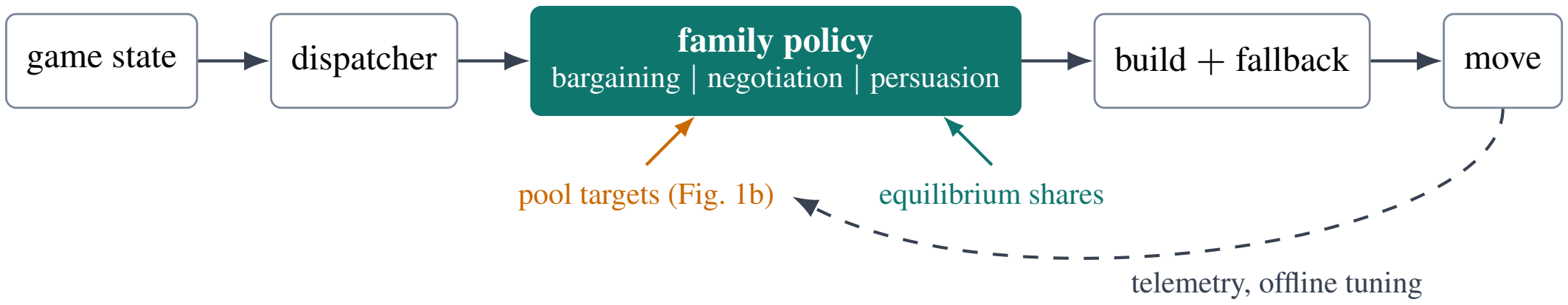
\fi

\end{document}